\documentclass[runningheads]{llncs}
\usepackage{graphicx}
\usepackage{amsmath,amssymb} 
\usepackage{color}

\usepackage{subcaption}
\usepackage{color}

\usepackage{times}
\usepackage{epsfig}
\usepackage{tabularx}
\usepackage[inline]{enumitem}
\usepackage{multirow}
\usepackage[dvipsnames]{xcolor}
\usepackage{booktabs}
\usepackage{algpseudocode,algorithm,algorithmicx}
\usepackage{xspace}
\usepackage{textcomp}

\usepackage[font=small,labelfont=bf]{caption} 

\usepackage{booktabs,siunitx}

\graphicspath{{figures/}}

\newcommand{\B}[1]{{\textbf{#1}}}
\newcommand{\SC}[1]{{\textsc{#1}}}
\newcommand{\commentout}[1]{}

\newcommand{\ie}{\textit{i}.\textit{e}.,~}
\newcommand{\eg}{\textit{e}.\textit{g}.,~}

\newcommand{\reffig}[1]{Figure~\ref{#1}}
\newcommand{\reftbl}[1]{Table~\ref{#1}}
\newcommand{\refsec}[1]{Section~\ref{#1}}

\newcommand\blfootnote[1]{%
  \begingroup
  \renewcommand\thefootnote{}\footnote{#1}%
  \addtocounter{footnote}{-1}%
  \endgroup
}

\usepackage{breakcites}
\definecolor{citecolor}{RGB}{34, 139, 34}
\usepackage[pagebackref=true,breaklinks=true,letterpaper=true,colorlinks,citecolor=citecolor, bookmarks=false]{hyperref}

\begin{document}
\title{Attributes as Operators: \\
Factorizing  Unseen Attribute-Object Compositions}

\titlerunning{Attributes as Operators}
%
\author{
	Tushar Nagarajan\inst{1} \and 
	Kristen Grauman\inst{2} 
}
%
\authorrunning{T. Nagarajan and K. Grauman}
%

\institute{
  The University of Texas at Austin \and Facebook AI Research \\
  \email{tushar@cs.utexas.edu, grauman@fb.com$^*$}
}

\maketitle              
\begin{abstract}
\blfootnote{*\emph{On leave from University of Texas at Austin (\texttt{grauman@cs.utexas.edu}).}}
We present a new approach to modeling visual attributes.  Prior work casts attributes in a similar role as objects, learning a latent representation where properties (e.g., \emph{sliced}) are recognized by classifiers much in the way objects (e.g., \emph{apple})  are. 
However, this common approach fails to separate the attributes observed during training from the objects with which they are composed, making it ineffectual when encountering new attribute-object compositions.  Instead, we propose to model attributes as \emph{operators}. Our approach learns a semantic embedding that explicitly factors out attributes from their accompanying objects, and also benefits from novel regularizers expressing attribute operators' effects (e.g., \emph{blunt} should undo the effects of \emph{sharp}).  
Not only does our approach align conceptually with the linguistic role of attributes as modifiers, but it also generalizes to recognize unseen compositions of objects and attributes. We validate our approach on two challenging datasets and demonstrate significant improvements over the state of the art. In addition, we show that not only can our model recognize unseen compositions robustly in an open-world setting, it can also generalize to compositions where objects themselves were unseen during training.
\end{abstract}
\section{Introduction}

Attributes are semantic descriptions that convey an object's properties---such as its materials, colors, patterns, styles, expressions, parts, or functions.   Attributes have proven to be an effective representation for faces and people~\cite{facetracer,relative-attributes,siddiquie-feris-cvpr2011,walk-learn-cvpr2016,face-attributes-iccv2015,yongjae-stn,lu-feris-cvpr2017}, catalog products~\cite{tamara-eccv2010,whittle-search,yu2017semantic,kimberly-iccv2017}, and generic objects and scenes~\cite{lampert-cvpr2009,farhadi-cvpr2009,Laffont14,sun-attributes-patterson,elhoseiny-cvpr2015,ziad-cvpr2016}.
Because they are expressed in natural language, attributes facilitate human-machine communication about visual content,  \eg for applications in image search~\cite{facetracer,whittle-search}, zero-shot learning~\cite{ziad-cvpr2016}, narration~\cite{baby-talk}, or image generation~\cite{yan2016attribute2image}.

Attributes and objects are fundamentally different entities: objects are physical things (nouns), whereas attributes are properties of those things (adjectives). Despite this fact, existing methods for attributes largely proceed in the same manner as state-of-the-art object recognition methods.  Namely, image examples labeled according to the attributes present are used to train discriminative models, \eg with a convolutional neural network~\cite{walk-learn-cvpr2016,face-attributes-iccv2015,yongjae-stn,lu-feris-cvpr2017,yu2017semantic,su-eccv2016}.

\begin{figure}[t!]
\centering
\includegraphics[width=\columnwidth]{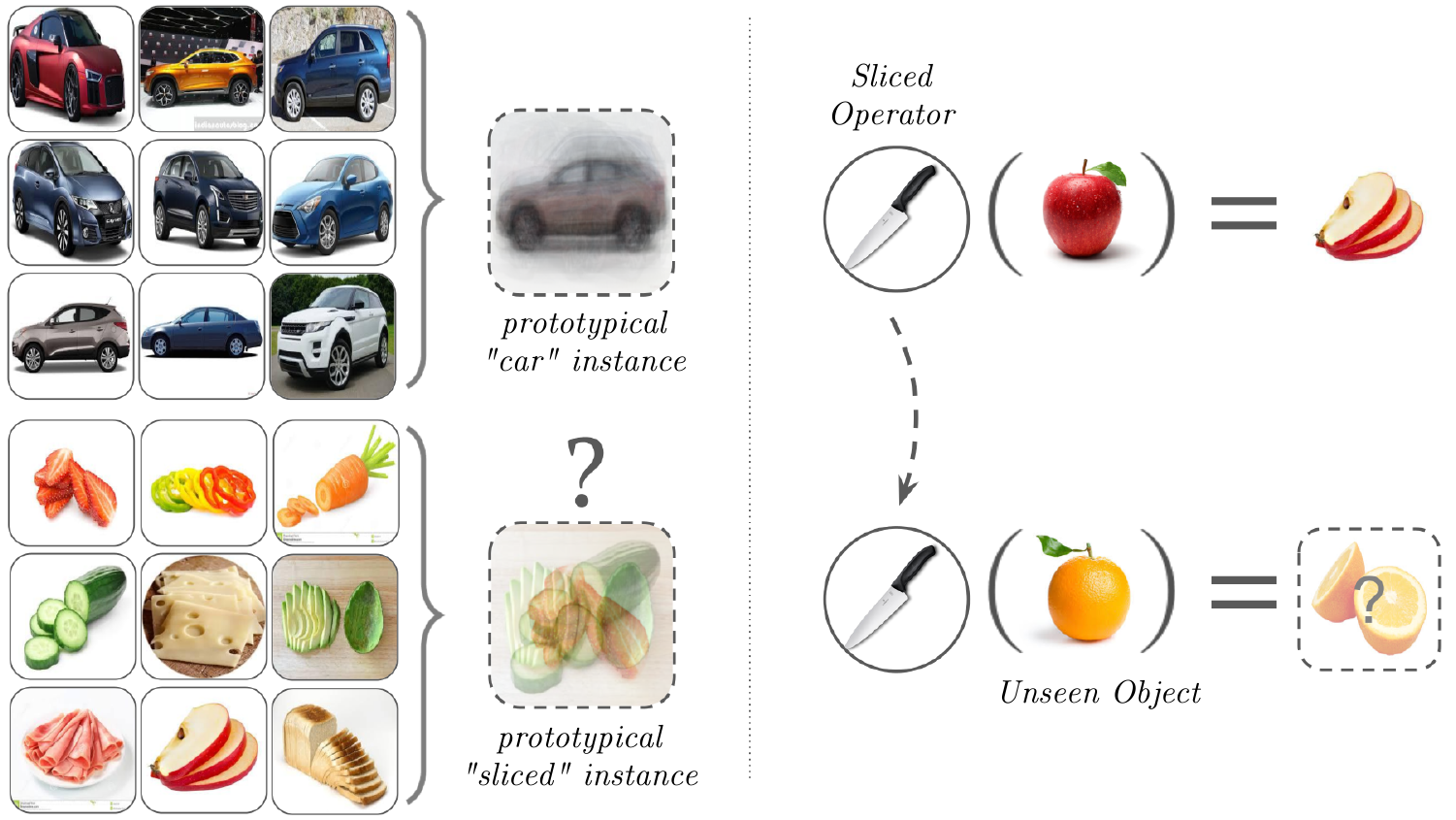}
\vspace{-0.2in}
\caption{ \textbf{Conceptual overview of our idea}. Left: 
Unlike for objects, it is difficult to learn a predictable visual prototype for an attribute (\eg ``sliced" as shown here).  Furthermore, standard visual recognition pipelines are prone to overfit to those object-attribute pairings observed during training.
Right: We propose to model attributes as operators, learning how they \emph{transform objects} rather than what they themselves look like. Once learned, the effects of the attribute operators are generalizable to new, unseen object categories.}
\label{fig:teaser}
\vspace{-0.2in}
\end{figure}

The latent vector encoding learned by such models is expected to capture an \emph{object-agnostic} attribute representation. Yet, achieving this is problematic, both in terms of data efficiency and generalization. Specifically, it assumes during training that 1) the attribute has been observed in combination with all potential objects (unrealistic and not scalable), and/or 2) an attribute's influence is manifested similarly across all objects (rarely the case, \eg ``old" influences church and shoe differently). We observe that with the attribute's meaning so intrinsically tied to the object it describes, an ideal attribute vector encoding may not exist.  See Figure~\ref{fig:teaser}, left.

In light of these issues, we propose to model attributes as \emph{operators} --- with the goal of learning a model \emph{for attribute-object composition itself} capable of explicitly factoring out the attributes' effect from their accompanying object representations.

First, rather than encode an attribute as a point in some embedding space, we encode it as a (learned) transformation that, when applied to an object encoding, modifies it to appropriately transform its appearance (see Figure~\ref{fig:teaser}, right).  
In particular, we formulate an embedding objective where compositions and images project into the same semantic space, allowing recognition of unseen attribute-object pairings in novel images.\footnote{We stress that this differs from traditional zero-shot object recognition~\cite{lampert-cvpr2009,dinesh-nips2014,ziad-cvpr2016}, where an \emph{unseen object} is defined by its (previously learned and class-agnostic) attributes.  In our case, we have \emph{unseen compositions} of objects and attributes.}

Second, we introduce novel regularizers during training that capitalize on the attribute-as-operator concept.  For example, one regularizer requires that the effect of applying an attribute and then its antonym to an object should produce minimal change in the object encoding (e.g., \emph{blunt} should ``undo" the effects of \emph{sharp}); another requires commutativity when pairs of attributes modify an object (e.g., a \emph{sliced red} apple is equivalent to a \emph{red sliced} apple). 

We validate our approach on two challenging datasets: MIT-States~\cite{misra2017red} and UT-Zappos~\cite{yu2014fine}.  Together, they span hundreds of objects, attributes, and compositions.
The results demonstrate the advantages of attributes as operators, in terms of the accuracy in recognizing unseen attribute-object compositions.  We observe significant improvements over state-of-the-art methods for this task~\cite{chen2014inferring,misra2017red}, with absolute improvements of 3\%-12\%. 
Finally, we show that our method is similarly robust whether identifying unseen compositions on their own or in the company of seen compositions---which is of great practical value for recognition in realistic, open world settings.
\vspace*{-0.1in}
\section{Related Work}

\noindent\textbf{Visual attributes}. 
Early work on visual attributes~\cite{facetracer,lampert-cvpr2009,farhadi-cvpr2009,relative-attributes} established the task of inferring mid-level semantic descriptions from images.  The research community has since explored many applications for attributes, including image search~\cite{facetracer,whittle-search,siddiquie-feris-cvpr2011}, zero-shot object categorization~\cite{lampert-cvpr2009,dinesh-nips2014,ziad-cvpr2016}, sentence generation~\cite{baby-talk} and fashion image analysis~\cite{tamara-eccv2010,huang-iccv2015,kimberly-iccv2017}.  Throughout, the standard approach to learn attributes is very similar to that used to learn object categories: discriminative classifiers with labeled examples.  In particular, today's best accuracies are obtained by training a deep convolutional neural network to classify attributes~\cite{walk-learn-cvpr2016,face-attributes-iccv2015,yongjae-stn,lu-feris-cvpr2017,su-eccv2016}.  Multi-task attribute training methods account for correlations between different attributes~\cite{lu-feris-cvpr2017,dinesh-cvpr2014,elhoseiny-cvpr2015,siddiquie-feris-cvpr2011}.  Our approach is a fundamental departure from all of the above: rather than consider attribute instances as points in some high-dimensional space that can be classified, we consider attributes as \emph{operators} that transform visual data from one condition to another.

\vspace{0.05in}
\noindent \textbf{Composition in language and vision}.  
In natural language processing, the composition of adjectives and nouns is modeled as single compositions~\cite{guevara2010regression,mitchell2008vector} or transformations (\ie an adjective transformation applied to the noun vector)~\cite{baroni2010nouns,socher2013recursive}.  Bridging such linguistic concepts to visual data, some work explores the correlation between similarity scores for color-object pairs in the language and visual domains~\cite{nguyen2014coloring}. 

Composition in vision has been studied in the context of modeling compound objects \cite{pezzelle2016building} (clipboard = clip + board), verb-object interactions \cite{sadeghi2011recognition,zhang2017visual} (riding a horse = person + riding + horse), and adjective-noun combinations~\cite{chen2014inferring,misra2017red,cruz2018neural} (fluffy towel = towel modified by fluffy).
All these approaches leverage the key insight that the characteristics of the composed entities could be very different from their constituents; however, they all subscribe to the traditional notion of representing constituents as vectors, and compositions as black-box modifications of these vectors. Instead, we model compositions as unique operators conditioned on the constituents (\eg for attribute-object composition, a different modification for each attribute).

Limited prior work on attribute-object compositions considers \emph{unseen compositions}, that is, where each constituent is seen during training, but new unseen compositions are seen at test time~\cite{chen2014inferring,misra2017red}.  Both methods construct classifiers for composite concepts using pre-trained linear classifiers for the ``seen'' primitive concepts, either with tensor completion~\cite{chen2014inferring} or neural networks~\cite{misra2017red}.  Recent work extends this notion to expressions connected by logical operators~\cite{cruz2018neural}.  We tackle unseen compositions as well. However, rather than treat attributes and objects alike as classifier vectors and place the burden of learning on a single network, we propose a factored representation of the constituents, modeling attribute-object composition as an attribute-specific invertible \emph{transformation} on object vectors. Our formulation also enables novel regularizers based on the attributes' linguistic meaning. Our model naturally extends to compositions where the objects themselves are unseen during training, unlike \cite{chen2014inferring,misra2017red} which requires an SVM classifier to be trained for every new object.
In addition, rather than exclusively predict unseen compositions as in~\cite{misra2017red}, we also study the more realistic scenario where \emph{all} compositions are candidates for recognition.

\vspace{0.05in}
\noindent\textbf{Visual transformations}.
The notion of visual ``states" has been explored from several  angles. 
Given a collection of images~\cite{isola2015discovering} or time-lapse videos~\cite{zhou2016learning,Laffont14}, methods can discover transformations that map between object states in order to create new images or visualize their relationships.  Given video input, action recognition can be posed as learning the visual state transformation, \eg how a person manipulates an object~\cite{fathi2013modeling,alayrac2017joint} or how activity preconditions map to postconditions~\cite{wang2016actions}.  
Given a camera transformation, other methods visualize the scene from the specified new viewpoint~\cite{jayaraman2015learning,zhou2016view}. While we share the general concept of capturing a visual transformation, we are the first to propose modeling attributes as operators that alter an object's state, with the goal of recognizing unseen compositions. 

\vspace{0.05in}
\noindent\textbf{Low-shot learning with sample synthesis}.
Recent work explores ways to generate synthetic training examples for classes that rarely occur, either in terms of features~\cite{dixit2017aga,hariharan2017low,lu2017zero,xian2017feature,zhu2017imagine} or entire images~\cite{yu2017semantic,choe2017face}. One part of our novel regularization approach also involves hypothetical attribute-transformed examples.  However, whereas prior work explicitly generates samples offline to augment the dataset, our feature generation  is an implicit process to regularize learning and works in concert with other novel constraints like inverse consistency or commutativity (see \refsec{sec:loss_function}).
\vspace*{-0.1in}
\section{Approach}

Our goal is to identify attribute-object compositions (\eg sliced banana, fluffy dog) in an image.  
Conventional classification approaches suffer from the long-tailed distribution of complex concepts \cite{sadeghi2011recognition,lu2016visual} and a limited capacity to generalize to unseen concepts.  
Instead, we model the composition process itself. We factorize out the underlying primitive concepts (attributes and objects) seen during training, and use them as building blocks to identify unseen combinations  during inference. Our approach is driven by the fundamental narrative: \emph{if we've seen a sliced orange, a sliced banana, and a rotten banana, can we anticipate what a rotten orange looks like?}

We model the composition process around the functional role of attributes. Rather than treat objects and attributes equally as vectors, we model attributes as invertible operators, and composition as an attribute-conditioned transformation \emph{applied} to object vectors.
Our recognition task then turns into an embedding learning task, where we project images and compositions into a common semantic space to identify the composition present. We guide the learning with novel regularizers that are consistent with the linguistic behavior of attributes.

In the following, we start by formally describing the embedding learning problem in \refsec{sec:manifold_recap}. 
We then describe the details of our embedding scheme for attributes and objects in \refsec{sec:attr_embedding}.
We present our optimization objective and auxiliary loss terms in \refsec{sec:loss_function}. Finally, we describe our training methodology in \refsec{sec:training_methodology}.

\begin{figure}[t!]
\centering
\begin{subfigure}[b]{\textwidth}
   \includegraphics[width=\columnwidth]{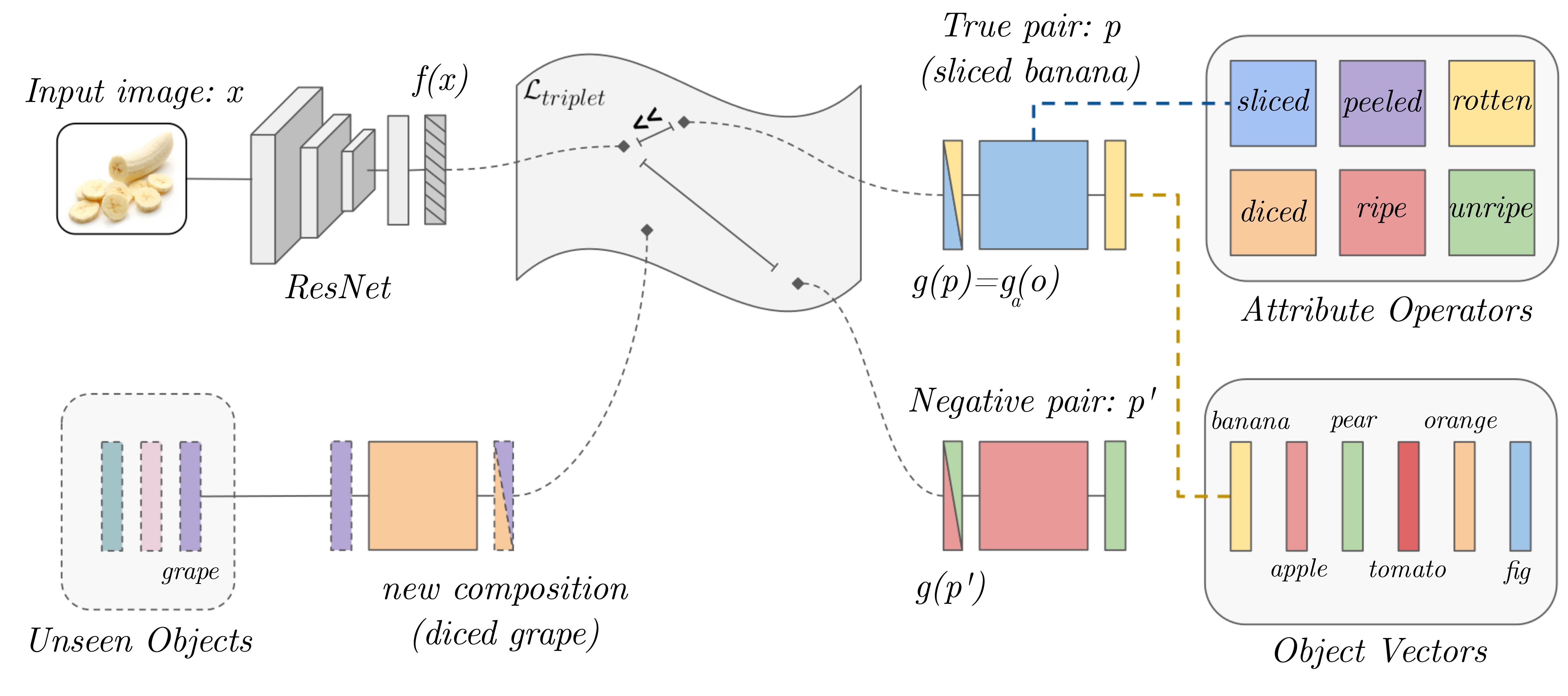}\vspace*{-0.1in}
   \caption{\textbf{Proposed model}. We propose a factorized model for attribute-object composition where objects are vectors (\eg GloVe~\cite{pennington2014glove} vectors, bottom right), attributes are operators (top right matrices), and composition is an attribute-specific transformation of an object vector ($g(p)$). We embed images $x$ and compositions $p$ in a space where distances represent compatibility between them (center). Because of the way compositions are factorized, known attributes may be \emph{assembled} with unseen objects, allowing our model to recognize new, unseen compositions in images (bottom left). Note that here \emph{object vectors} are category-level embeddings, not images.}
   \label{fig:model} \vspace*{0.15in}
\end{subfigure}
\begin{subfigure}[b]{\textwidth}
   \includegraphics[width=\columnwidth]{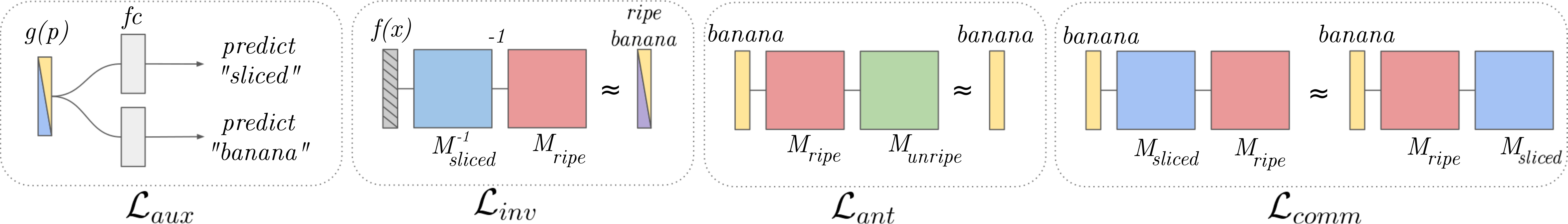}
   \caption{\textbf{Proposed regularizers}. We propose several regularizers that conform with the linguistic meaning of attributes. $\mathcal{L}_{aux}$ ensures that the identity of the attribute/object is not lost during composition; $\mathcal{L}_{inv}$ \emph{swaps out} attributes to implicitly synthesize new compositions for training;  $\mathcal{L}_{ant}$ models antonyms (``unripe" should \emph{undo} the effects of ``ripe"); and $\mathcal{L}_{comm}$ models the commutative property of attributes (a \emph{ripe sliced} banana is the same as a \emph{sliced ripe} banana).}
   \label{fig:regularizers}
\end{subfigure}
\vspace{-0.2in}
\caption{\textbf{Overview of proposed approach.}  Best viewed in color}
\end{figure}

\subsection{Unseen pair recognition as embedding learning} \label{sec:manifold_recap}
\vspace*{-0.05in}
We train a model that learns a mapping from a set of images $\mathcal{X}$ to a set of attribute-object pairs $\mathcal{P} = \mathcal{A} \times \mathcal{O}$.  For example, ``old-dog'' is one attribute-object pairing.  We divide the set of pairs into two disjoint sets: $\mathcal{P}_s$, which is a set of pairs that is seen during training and is used to learn a factored composition model, and $\mathcal{P}_u$, which is a set of pairs unseen during training, yet perfectly valid to encounter at test time.  While $\mathcal{P}_s$ and $\mathcal{P}_u$ are completely disjoint, their constituent attributes and objects are observed in some (other) composition during training. Our images contain objects with a single attribute label associated with them, \ie each image has a unique pair label $p \in \mathcal{P}$.

During training, given an image $x \in \mathcal{X}$ and its corresponding pair label $p \in \mathcal{P}_s$, we learn two embedding functions $f(x)$ and $g(p)$ to project them into a \emph{common semantic space}.  
For $f(x)$, we use a pretrained ResNet18 \cite{he2016deep} followed by a linear layer. For $g(p)$, we introduce an attribute-operator model, described in detail in \refsec{sec:attr_embedding}.

We learn the embedding functions such that in this space, the Euclidean distance between the image embedding $f(x)$ and the correct pair embedding $g(p)$ is minimized, while the distance to all incorrect pairs is maximized.
Distance in this space represents \emph{compatibility}---\ie a low distance between an image and pair embedding implies the pair is present in the image.  Critically, once $g(p)$ is learned, even an unseen pair can be projected in this semantic space, and its compatibility with an image can be assessed.  See \reffig{fig:model}.

During inference, we compute and store the pair embeddings of \emph{all}  potential pair candidates from $\mathcal{P}$ using our previously learned composition function $g(.)$. When presented with a new image, we embed it as usual using $f(.)$, and identify which of the pair embeddings is closest to it. Note how $\mathcal{P}$ includes both pairs seen in training as well as unseen attribute-object compositions; 
recognizing the latter would not be possible if we were doing a simple classification among the previously seen combinations.

\vspace*{-0.1in}
\subsection{Attribute-operator model for composition} \label{sec:attr_embedding}

As discussed above, the conventional approach treats attributes much like objects, both occupying some point/region in an embedding space~\cite{siddiquie-feris-cvpr2011,walk-learn-cvpr2016,face-attributes-iccv2015,yongjae-stn,lu-feris-cvpr2017,su-eccv2016,dinesh-cvpr2014,elhoseiny-cvpr2015}.

On the one hand, it is meaningful to conjure a latent representation for an ``attribute-free object''---for example, \emph{dog} exists as a concept before we specialize it to be a spotted or fluffy dog.  In fact, in the psychology of perception, one way to characterize a so-called basic-level category is by its affordance of a single mental prototype~\cite{rosch1976basic}.  
On the other hand, however, it is problematic to conjure an ``object-free attribute''.  What does it mean to map ``fluffy'' as a concept in a semantic embedding space?  What is the visual prototype of ``fluffy"?  See Figure~\ref{fig:teaser}.

We contend that a more natural way of describing attributes is in how they \emph{modify} the objects they refer to. Images of a ``dog" and a ``fluffy dog" help us estimate what the concept ``fluffy" refers to. Moreover, these modifications are strongly conditioned on the object they describe (``fluffy" exhibits itself significantly differently in ``fluffy dog" compared to ``fluffy pillow").
In this sense, attribute behavior bears some resemblance to geometric transformations.  For example, rotation can be perfectly represented as an orthogonal matrix acting on a vector. Representing rotation as a vector, and its action as some additional function, would be needlessly complicated and unintuitive. 

With this in mind, we represent each object category $o \in \mathcal{O}$ as a $D$-dimensional vector, which denotes a prototypical object instance. Specifically, we use GloVe word embeddings~\cite{pennington2014glove} for the object vector space.  Each attribute $a \in \mathcal{A}$ is a parametrized function $g_a: \mathcal{R}^{D} \rightarrow \mathcal{R}^{D}$ that modifies an object representation to exhibit that attribute, and brings it to the semantic space where images reside. For simplicity, we consider a linear transform for $g_a$, represented by a $D \times D$ matrix $M_a$:
\begin{equation}
g(p) = g_a(o) = M_a o,
\end{equation}
though the proposed framework (excluding the inverse consistency regularizer)  naturally supports more complex functions for $g_a$ as well. See Figure~\ref{fig:model}, top right. 

Interesting properties arise from our attribute-operator design.
First, factorizing composition as a matrix-vector product facilitates transfer: an unseen pair can be represented by applying a learned attribute operator to an appropriate object vector (Figure~\ref{fig:model}, bottom left).
Secondly, since images and compositions reside in the same space, it is possible to \emph{remove} attributes from an image by applying the inverse of the transformation; multiple attributes can be applied consecutively to images; and the structure of the attribute space can be coded into how the transformations behave. Below we discuss how we leverage these properties to regularize the learning process (Sec.~\ref{sec:loss_function}).

\subsection{Learning objective for attributes as operators} \label{sec:loss_function}

Our training set consists of $n$ images and their pair labels, $\{(x_1,p_1),\dots,(x_n,p_n)\}$.  
We design a loss function to efficiently learn to project images and composition pairs  to a common embedding space. We begin with a standard triplet loss. The loss for an image $x$ with pair label $p=(a,o)$ is given by:
\begin{equation}
\mathcal{L}_{triplet} =  \max \left ( 0, d(f(x), M_ao) - d(f(x), M_{a^\prime}o^\prime) + m \right ),
\forall\ a^\prime \neq a \lor o^\prime \neq o,
\label{eq:triplet}
\end{equation}
where $d$ denotes Euclidean distance, and $m$ is the margin value, which we keep fixed at 0.5 for all our experiments. In other words, the embedded image ought to be closer to its object transformed by the specified attribute $a$ than other attribute-object pairings.

Thus far, the loss is similar in spirit to embedding based zero-shot learning methods \cite{xian2017zero}, and more generally to triplet-loss based representation learning methods \cite{cheng2016person,hoffer2015deep,schroff2015facenet}. We emphasize that our focus is on learning a model for the composition operation; a triplet-loss based embedding is merely an appropriate framework that facilitates this. In the following, we extend this framework to effectively accommodate attributes as operators and inject our novel linguistic-based regularizers.

\vspace{0.05in}
\noindent\textbf{Object and attribute auxiliaries}.
In our model, both the attribute operator and object vector, and thereby their composition, are learnable parameters. It is possible that one element of the composition (either attributes or objects) will dominate during optimization, and try to capture all the information instead of learning a factorized model. This could lead to a composition representation, where one component does not adequately feature. To address this, we introduce an auxiliary loss term that forces the composed representation to be discriminative, \ie it must be able to predict both the attribute and object involved in the composition:
\begin{equation}
\mathcal{L}_{aux} = - \sum_{i \in \mathcal{A}} \delta_{ai}\ log(p_{a}^{i}) - \sum_{i \in \mathcal{O}} \delta_{oi}\ log(p_{o}^{i}),
\end{equation}
where $\delta_{yi}=1$ iff $y=i$, and $p_a$ and $p_o$ are the outputs of softmax linear classifiers trained to discriminate the attributes and objects, respectively. 
This auxiliary supervision ensures that the identity of the attribute and the object are not lost in the composed representation---in effect, strongly incentivizing a factorized representation.

\vspace{0.05in}
\noindent\textbf{Inverse consistency}.
We exploit the invertible nature of our attributes to implicitly synthesize new training instances to regularize our model further. More specifically, we \emph{swap out} an actual attribute $a$ from the training example for a randomly selected one $a^\prime$, and construct another triplet loss term to account for the new composition:
\begin{equation}
\begin{split}
f(x^\prime) & := M_{a^\prime}M_a^{-1}f(x) \\
\mathcal{L}_{inv} &= \max\left ( 0, d(f(x^\prime), M_{a^\prime}o) - d(f(x^\prime), M_{a}o) + m\right ),
\end{split}
\end{equation}
where the triplet loss notation is in the same form as Eq~\ref{eq:triplet}.

Here $M_{a^\prime}M_a^{-1}$ represents the removal of attribute $a$ to arrive at the ``prototype object" description of an image, and then the application of attribute $a^\prime$ to imbue the object with a new attribute. As a result, $f(x^\prime)$ represents a pseudo-instance with a new attribute-object pair, helping the model generalize better. 

The pseudo-instances generated here are inherently noisy, and factoring them in directly (as a new instance) may obstruct training. To mitigate this, we select our negative example to target the more direct, and thus simpler consequence of this swapping. For example, when we swap out ``sliced" for ``ripe" from a \emph{sliced banana} to make a \emph{ripe banana}, we focus on the more obvious fact---that it is no longer ``sliced"---by picking the original composition (\emph{sliced banana}) as the negative, rather than sampling a completely new one.

\vspace{0.05in}
\noindent\textbf{Commutative attribute operators}. 
Next we constrain the attributes to respect the commutative property.  For example,  applying the ``sliced" operator after the ``ripe" operator is the same as applying ``ripe" after ``sliced", or in other words a \emph{ripe sliced} banana is the same as a \emph{sliced ripe} banana.  This commutative loss is expressed as: 
\begin{align}
\mathcal{L}_{comm} &= \sum_{a, b \in \mathcal{A}} \left \| M_a(M_bo) - M_b(M_ao) \right \|_2.
\end{align}
This loss forces the attribute transformations to respect the notion of \emph{attribute composability} we observe in the context of language.

\vspace{0.05in}
\noindent\textbf{Antonym consistency}. 
The final linguistic structure of attributes we aim to exploit is antonyms. For example, we hypothesize that the ``blunt" operator should \emph{undo} the effects of the ``sharp" operator.  To that end, we consider a loss term that operates over pairs of antonym attributes $(a,a^\prime)$:\vspace*{-0.1in}
\begin{align}
\mathcal{L}_{ant} &= \sum_{a, a^\prime \in \mathcal{A}} \left \| M_{a^\prime}(M_ao) - o \ \right \|_2.
\end{align}
For the MIT-States dataset (cf.~Sec.~\ref{sec:results}), we manually identify 30 antonym pairs like ancient/modern, bent/straight, blunt/sharp.
Figure~\ref{fig:regularizers} recaps all the regularizers.

\subsection{Training and inference} \label{sec:training_methodology}

We minimize the combined loss function ($\mathcal{L}_{triplet}+\mathcal{L}_{aux} + \mathcal{L}_{inv} + \mathcal{L}_{comm} + \mathcal{L}_{ant}$) over all the training images, and train our network end to end. The learnable parameters are: the linear layer for $f(x)$, the matrices for every attribute $M_a$, $\forall a \in \mathcal{A}$, the object vectors $\forall o \in \mathcal{O}$ and the two fully-connected layers for the auxiliary classifiers.

During training, we embed each labeled image $x$ in a semantic space using $f(x)$, and apply its attribute operator $g_a$ to its object vector $o$ to get a composed representation $g_a(o)$.  The triplet loss pushes these two representations close together, while pushing incorrect pair embeddings apart. Our regularizers further make sure compositions are discriminative; attributes obey the commutative property; they undo the effects of their antonyms; and we implicitly synthesize instances with new compositions. 

For inference, we compute and store the embeddings for all candidate pairs, $g_a(o)$, $\forall o \in \mathcal{O}$ and $\forall a \in \mathcal{A}$. When a new image $q$ arrives, we sort the pre-computed embeddings by their distance to the image embedding $f(q)$, and identify the compositions with the lowest distances. The distance calculations can be performed quickly on our dataset with a few thousand pairs. Intelligent pruning strategies may be employed to reduce the search space for larger attribute/object vocabularies. We stress that the novel image can be assigned to an unseen composition absent in training images.
We evaluate accuracy on the nearest composition $\hat{p_q} = (o_q, a_q)$ as our datasets support instances with single attributes. 
\vspace*{-0.1in}
\section{Experiments}\label{sec:results}

Our experiments explore the impact of modeling attributes as operators, particularly for recognizing unseen combinations of objects and attributes.

\vspace*{-0.1in}
\subsection{Experimental setup}

\noindent\textbf{Datasets}. We evaluate our method on two datasets:

\begin{itemize}
\item \textbf{MIT-States} \cite{isola2015discovering}: This dataset has 245 object classes, 115 attribute classes and
$\sim$53K images.  There is a wide range of objects (\eg \emph{fish}, \emph{persimmon}, \emph{room}) and attributes (\eg \emph{mossy}, \emph{deflated}, \emph{dirty}). On average, each object instance is modified by one of the 9 attributes it affords. We use the \emph{compositional} split described in \cite{misra2017red} for our experiments, resulting in disjoint sets of pairs---about 1.2K pairs in $\mathcal{P}_s$ for training and 700 pairs in $\mathcal{P}_u$ for testing. 
\vspace*{0.1in}

\item \textbf{UT-Zappos50k} \cite{yu2017semantic}: This dataset contains 50K images of shoes with attribute labels. We consider the subset of $\sim$33K images that contain annotations for material attributes of shoes (\eg \emph{leather}, \emph{sheepskin}, \emph{rubber}); see Supp.  
The object labels are shoe types (\eg \emph{high heel}, \emph{sandal}, \emph{sneaker}). 
We split the data randomly into disjoint sets, yielding 83 pairs in $\mathcal{P}_s$ for training and 33 pairs in $\mathcal{P}_u$ for testing, over 16 attribute classes and 12 object classes.
\end{itemize}

The datasets are complementary.  While MIT-States covers a wide array of everyday objects and attributes, UT-Zappos focuses on a fine-grained domain of shoes. In addition, object annotations in MIT-States are very sparse (some classes have just 4 images), while the UT-Zappos subset has at least 200 images per object class. 

\vspace{0.05in}
\noindent\textbf{Evaluation metrics}. We report top-1 accuracy on recognizing pair compositions. We report this accuracy in two forms: (1) Over only the unseen pairs, which we refer to as the \textbf{closed world setting}.  During test time, we compute the distance between our image embedding and only the pair embeddings of the unseen pairs $\mathcal{P}_u$, and select the nearest one. The closed world setting artificially reduces the pool of allowable labels at test time to \emph{only} the unseen pairs. This is the setting in which \cite{misra2017red} report their results.
(2) Over both seen and unseen pairs, which we call the \textbf{open world setting}.  During test time, we consider all pair embeddings in $\mathcal{P}$ as candidates for recognition. This is more realistic and challenging, since no assumptions are made about the compositions present. 
We aim for high accuracy in both these settings. We report the \emph{harmonic mean} of these accuracies given by $h \mbox{-} mean = 2 \ast (open \ast closed)/(open + closed)$, as a consolidated metric. 
Unlike the arithmetic mean, it penalizes large performance discrepancies between settings. 
The harmonic mean is recommended to handle a similar discrepancy between seen/unseen accuracies in ``generalized'' zero-shot learning~\cite{xian2017zero}, and is now widely adopted as an evaluation metric \cite{verma2018generalized,xian2018feature,chen2018zero,wang2017alternative}.

\vspace{0.05in}
\noindent\textbf{Implementation details}.
For all experiments, we use an ImageNet \cite{russakovsky2015imagenet} pretrained ResNet-18 \cite{he2016deep} for $f(x)$. For fair comparison, we do not finetune this network.
We project our images and compositions to a $D=300$-dim.~embedding space.
We initialize our object and attribute embeddings with GloVe~\cite{pennington2014glove} word vectors where applicable, and initialize attribute operators with the identity matrix as this leads to more stable training. All models are implemented in PyTorch. ADAM with learning rate $1e -4$ and batch size 512 is used. The attribute operators are trained with learning rate $1e-5$ as they encounter larger changes in gradient values. Our code is available at \href{https://github.com/Tushar-N/attributes-as-operators}{\texttt{github.com/attributes-as-operators}}.

\vspace{0.05in}
\noindent\textbf{Baselines and existing methods}. We compare to the following methods:
\begin{itemize}
\item \textbf{\SC{VisProd}} uses independent classifiers on the image features to predict the attribute and object.  It represents methods that do not explicitly model the composition operation. The probability of a pair is simply the product of the probability of each constituent: $P(a,o) = P(a)P(o)$. We report two versions, differing in the choice of the classifier used to generate the aforementioned probabilities: \SC{VisProd(SVM)} uses a Linear SVM (as used in \cite{misra2017red}), and \SC{VisProd(NN)} uses a single layer softmax regression model. 

\item \textbf{\SC{AnalogousAttr}}~\cite{chen2014inferring} trains a linear SVM classifier for each seen pair, then uses Bayesian Probabilistic Tensor Factorization (BPTF) to infer classifier weights for unseen compositions. We use the same existing code\footnote{\url{https://www.cs.cmu.edu/~lxiong/bptf/bptf.html}} as~\cite{chen2014inferring} to recreate this model.

\item \textbf{\SC{RedWine}}~\cite{misra2017red} trains a neural network to transform linear SVMs for the constituent concepts into classifier weights for an unseen combination. Since the authors' code was not available, we implement it  ourselves following the paper closely.  We train the SVMs with image features consistent with our models. We verify we could reproduce their results with VGG (network they employed), then upgrade its features to ResNet to be more competitive with our approach. 

\item \textbf{\SC{LabelEmbed}} is like the \SC{RedWine} model, except it composes word vector representations rather than classifier weights. We use pretrained GloVe~\cite{pennington2014glove} word embeddings. This is the LabelEmbed baseline designated in~\cite{misra2017red}. 

\item \textbf{\SC{LabelEmbed+}} is an improved version of \SC{LabelEmbed} where (1) We embed both the constituent inputs \emph{and} the image features using feed-forward networks into a semantic embedding space of dimension $D$, and (2) We allow the input representations to be optimized during training. See Supp.~for details.
\end{itemize} 

To our knowledge~\cite{chen2014inferring,misra2017red} are the most relevant methods for comparison, as they too address recognition of unseen object-attribute pairs.
For all methods, we use the same ResNet-18 image features used in our method; this ensures any performance differences can be attributed to the model rather than the CNN architecture. For all neural models, we ensure that the number of parameters and model capacity are similar to ours.

\begin{table*}[t]
\centering
\small
\begin{tabular}{l*{4}{S[table-format=3.2]}|*{4}{S[table-format=3.2]}}
                                            & \multicolumn{4}{c}{MIT-States}                    & \multicolumn{4}{c}{UT-Zappos}                     \\
\cline{2-5} \cline{6-9} 
                                            & {closed}   & {open}     & {+obj}     & {h-mean}   & {closed}   & {open}     & {+obj}     & {h-mean}   \\
\midrule
\SC{Chance}                                 & 0.1        & 0.05       & 0.9        & 0.1        & 3.0        & 0.9        & 6.3        & 1.3        \\
\SC{VisProd(SVM)}                           & 11.1       & 2.4        & 21.6       & 3.9        & 46.8       & 4.1        & 17.8       & 7.5        \\
\SC{VisProd(NN)}                            & 13.9       & 2.8        & 22.6       & 4.7        & \B{49.9}   & 4.8        & 18.1       & 8.8        \\
\SC{AnalogousAttr} \cite{chen2014inferring} & 1.4        & 0.2        & 22.4       & 0.4        & 18.3       & 3.5        & 16.9       & 5.9        \\
\SC{RedWine} \cite{misra2017red}            & 12.5       & 3.1        & 18.3       & 5.0        & 40.3       & 2.1        & 10.5       & 4.0        \\
\SC{LabelEmbed}                             & 13.4       & 3.3        & 18.8       & 5.3        & 25.8       & 5.2        & 11.1       & 8.7        \\
\SC{LabelEmbed+}                            & \B{14.8}   & 5.7        & 27.2       & 8.2        & 37.4       & 9.4        & 19.4       & 15.0       \\ 
\midrule
\SC{Ours}                                   & 12.0       & \B{11.4}   & \B{49.3}   & \B{11.7}   & 33.2       & \B{23.4}   & \B{38.3}   & \B{27.5}   \\
\bottomrule
\end{tabular}
\caption{\textbf{Accuracy (\%) on unseen pair detection.} Our method outperforms all previous methods in the open world setting.  It also is strongest in the consolidated harmonic mean (h-mean) metric that accounts for both the open and closed settings. Our method's gain is significantly wider when we eliminate the pressure caused by scarce object training data, by providing oracle object labels during inference to all methods (``+obj"). The harmonic mean is calculated over the open and closed settings only (it does not factor in +obj).}
\vspace*{-0.25in}
\label{tab:unseen_pairs}
\end{table*}

\subsection{Quantitative results: recognizing object-attribute compositions}

\noindent\textbf{Detecting unseen compositions}. 
\reftbl{tab:unseen_pairs} shows the results. Our method outperforms all previously reported results and baselines on both datasets by a large margin---around 6\% on MIT-States and 14\% on UT-Zappos in the open world setting---indicating that it learned a strong model for visual composition.

The absolute accuracies on the two datasets are fairly different. Compared to UT-Zappos, MIT-States is more difficult owing to a larger number of attributes, objects, and unseen pairs. Moreover, it has fewer training examples for primitive object concepts, leading to a lower accuracy overall.

Indeed, if an oracle provides the true object label on a test instance, the accuracies are much more consistent across both datasets  (``+obj'' in \reftbl{tab:unseen_pairs}). This essentially trims the search space down to the attribute afforded by the object in question, and serves as an upper bound for each method's accuracy. On MIT-States, without object labels, the gap between the strongest baseline and our method is about 6\%, which widens significantly to about 22\% when object labels are provided (to all methods). On UT-Zappos, all methods improve with the object oracle, yet the gap is more consistent with and without (14\% vs. 19\%).  This is consistent with the datasets' disparity in label distribution; the model on UT-Zappos learns a good object representation by itself. 

\SC{AnalogousAttr}~\cite{chen2014inferring} varies significantly between the two datasets; it relies on having a partially complete set of compositions in the form of a tensor, and uses that information to ``fill in the gaps". For UT-Zappos, this tensor is 43\% complete, making completion a relatively simpler task compared to MIT-States, where the tensor is only 4\% complete. We believe that over-fitting due to this extreme sparsity is the reason we observe low accuracies for \SC{AnalogousAttr} on this dataset.

In the closed world setting, our method does not perform as well as some of the other baselines.  However, this setting is contrived and arguably a weaker indication of model performance.  In the closed world, it is easy for a method to produce biased results due to the artificially pruned label space during inference. For example, the attribute ``young" occurs in only \emph{one} unseen composition during test time---``young iguana". Since all images during test time that contain iguanas \emph{are} of ``young iguanas", an attribute-blind model is also perfectly capable of classifying these instances correctly, giving a false sense of accuracy. In practical applications, the separation into seen and unseen pairs arises from natural data scarcity. In that setting, the ability to identify unseen compositions \emph{in the presence of known compositions}, \ie the open world, is a critical metric. 

The lower performance in the closed world appears to be a side-effect of preventing overfitting to the subset of closed-world compositions. All models except ours have a large difference between the closed and open world accuracy. Our model operates robustly in both settings, maintaining similar accuracies in each. Our model outperforms the other models in the harmonic mean metric as well by about 3\% and 12\% on MIT-States and UT-Zappos, respectively.

\begin{table}[t]
\centering
\small
\vspace*{-0.15in}
\begin{tabular}{l*{3}{S[table-format=3.2]}|*{3}{S[table-format=3.2]}}
                    & \multicolumn{3}{c}{MIT-States}      & \multicolumn{3}{c}{UT-Zappos}             \\ \cline{2-4} \cline{5-7} 
                    & {closed}   & {open}      & {h-mean} & {closed}   & {open}          & {h-mean}   \\ \midrule
\SC{Base}           & \B{14.2}   & 2.1         & 3.7      & \B{46.2}   & 13.1            & 20.4       \\
\SC{+inv}           & 14.0       & 2.7         & 4.5      & 45.7       & 14.2            & 21.7       \\
\SC{+aux}           & 10.3       & 9.5         & 9.9      & 33.2       & 26.5            & 29.5       \\
\SC{+aux+inv}       & 10.4       & 9.8         & 10.1     & 33.1       & 26.2            & 29.2       \\
\SC{+aux+comm}      & 11.4       & 10.8        & 11.1     & 38.1       & \B{29.7}        & \B{33.4}   \\
\SC{+aux+ant}       & 8.9        & 8.8         & 8.8      & {-}        & {-}             & {-}        \\ 
\SC{+aux+inv+comm}  & 12.0       & \B{11.4}    & \B{11.7} & 33.2       & 23.4            & 27.5       \\ \bottomrule
\end{tabular}
\caption{\textbf{Ablation study of regularizers used.} The auxiliary classifier loss is essential to our method. Adding other regularizers that are consistent with how attributes function also produces boosts in accuracy in most cases, highlighting the merit of thinking of \emph{attributes as operators}. }
\vspace*{-0.25in}
\label{tab:regularizers}
\end{table}

\vspace{0.05in}
\noindent\textbf{Effect of regularizers}. \reftbl{tab:regularizers} examines the effects of each proposed regularizer on the performance of our model. 
We see that the auxiliary classification loss stabilizes the learning process significantly, and results in a large increase in accuracy on both datasets. For MIT-States, including the inverse consistency and the commutative operator regularizers provide small boosts and a reasonable increase when used together. For UT-Zappos, the effect of inverse consistency is less pronounced, possibly because the abundance of object training data makes it redundant. The commutative regularizer provides the biggest improvement of 4\%. Antonym consistency is not very helpful on MIT-States, perhaps due to the wide visual differences between some antonyms. For example, ``ripe" and ``unripe" for fruits produce vibrant color changes, and \emph{undoing} one color change does not directly translate to \emph{applying} the other \ie ``ripe" may not be the \emph{visual inverse} of ``unripe".\footnote{Attributes for UT-Zappos are centered around materials of shoes (\emph{leather}, \emph{cotton}) and so lack antonyms, preventing us from experimenting with that regularizer.} These ablation experiments show the merits of pushing our model to be consistent with how attributes operate.

Overall, the results on two challenging and diverse datasets strongly support our idea to model attributes as operators.  Our method consistently outperforms state-of-the-art methods. Furthermore, we see the promise of injecting novel linguistic/semantic operations into attribute learning.  

\begin{figure}[t]
\centering
\includegraphics[width=\columnwidth]{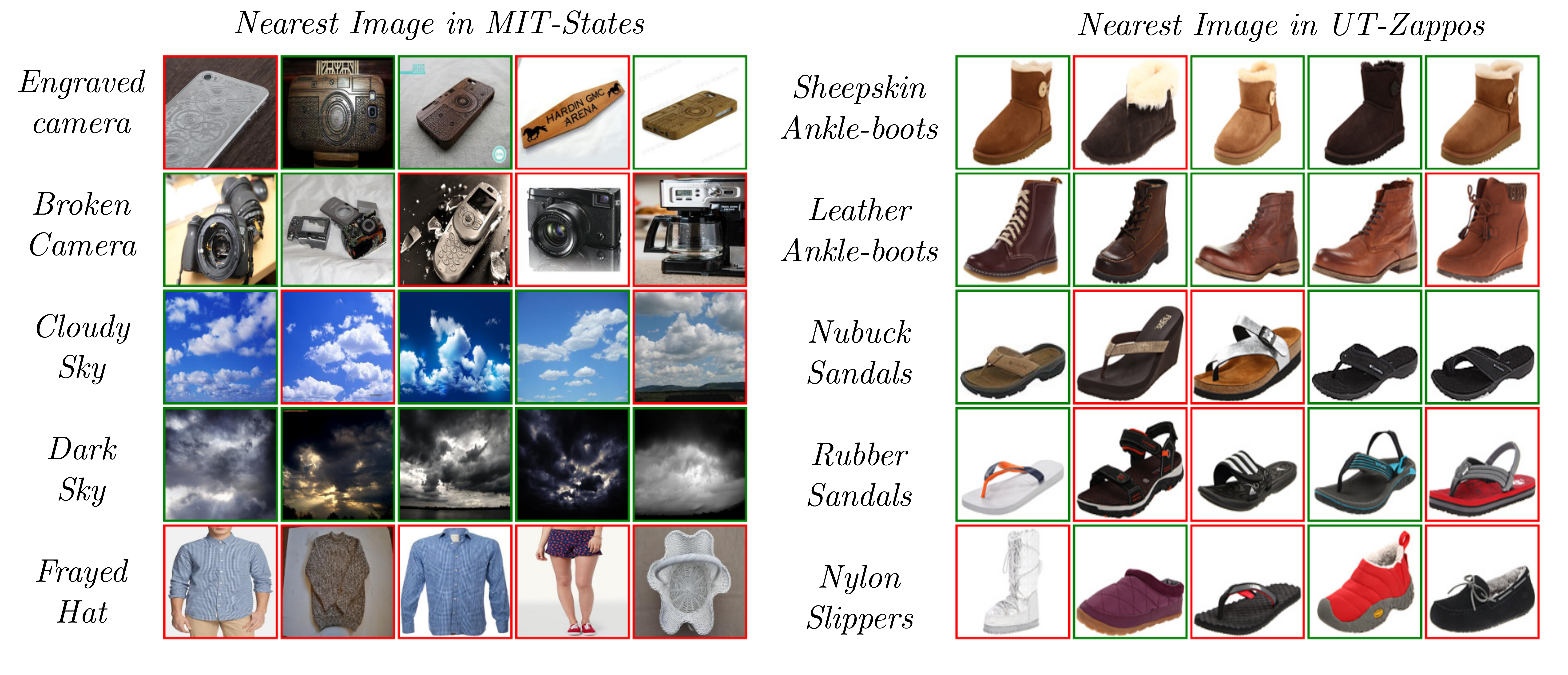}
\vspace{-0.25in}
\caption{ \textbf{Top retrieval results for unseen compositions}. Unseen compositions are posed as textual queries on MIT-States (left) and UT-Zappos (right). These attribute-object pairs are completely unseen during training; the representation for them is generated using our factored composition model. 
We highlight correctly retrieved instances with a green border, and incorrect ones with red. Last row shows failure cases.} 
\label{fig:retrival-unseen}
\vspace{-0.2in}
\end{figure}

\subsection{Qualitative results: retrieving images for unseen descriptions}

Next, we show examples of our approach at work to recognize unseen compositions.

\vspace{0.05in}
\noindent\textbf{Image retrieval for unseen compositions.} With a learned composition model in place, our method can retrieve relevant images for textual queries for object-attribute pairs unseen during training. The query itself is in the form of an attribute $a$ and an object $o$; we embed them, and all the image candidates $x$, in our semantic space, and select the ones that are nearest to our desired composition. We stress that these compositions are completely new and arise from our model's factored representation of composition. 

\reffig{fig:retrival-unseen} shows examples. The query is shown in text, and the top 5 nearest images in embedding space are shown alongside. Our method accurately distinguishes between attribute ``states" of the same object to retrieve relevant images for the query. The last row shows failure cases. We observe characteristic failures for compositions involving some under-represented object classes in training pairs. For example, compositions involving ``hat'' are poorly learned as it features in only two training compositions. We also observe common failures involving ambiguous labels (examples of \emph{moldy bread} are also often \emph{sliced} in the data). 

\begin{figure}[t]
\centering
\includegraphics[width=\columnwidth]{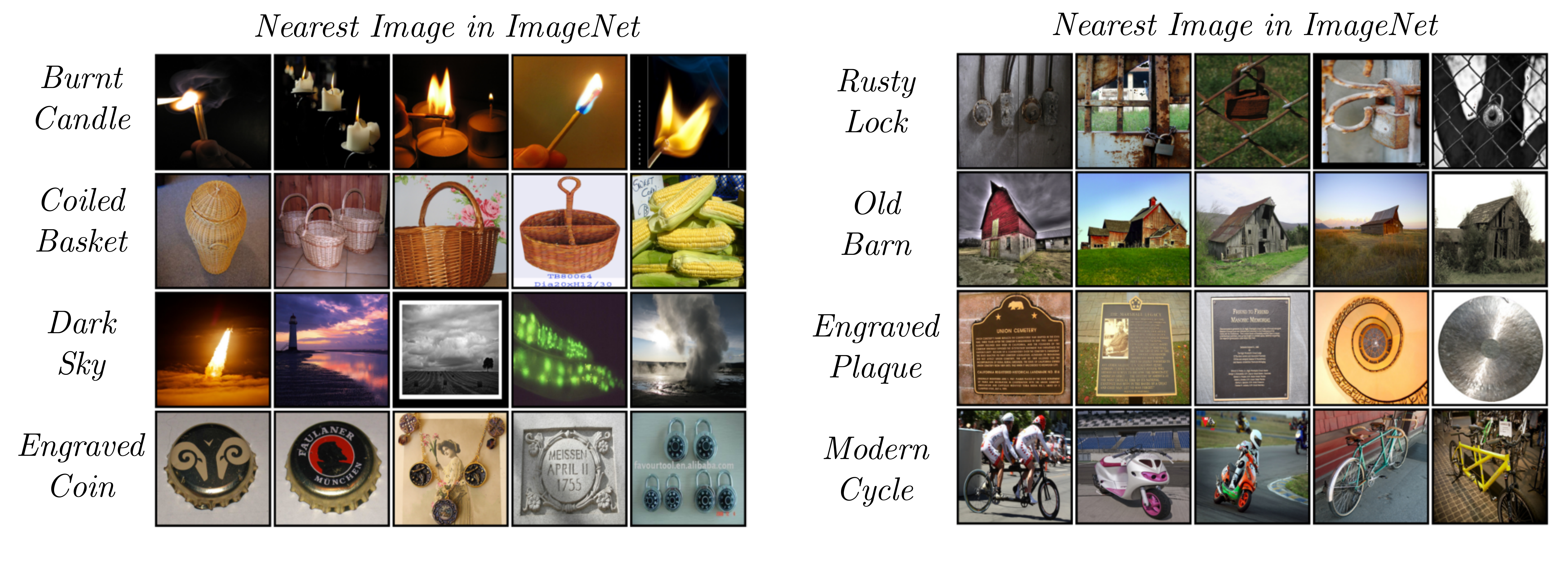}
\vspace{-0.2in}
\caption{\textbf{Top retrieval results in the out-of-domain setting}. Images are retrieved from an unseen domain, ImageNet. Left: Our method can successfully retrieve unseen compositions from images in the wild. Right: Retrievals on out-of-domain compositions. Compositions involving objects that are not even present in our dataset (like lock and barn) can be retrieved using our model's factorized representation.}
\label{fig:retrieval-unseen-unseen}
\vspace{-0.2in}
\end{figure}

\vspace{0.05in}
\noindent\textbf{Image retrieval for out-of-domain compositions}. 
\reffig{fig:retrieval-unseen-unseen} takes this task two steps further.  First, we perform retrieval on an image database disjoint from training to demonstrate robustness to domain shift in the open world setting.
\reffig{fig:retrieval-unseen-unseen} (left) shows retrievals from the ImageNet validation set, a set of 50K images disjoint from MIT-States. 
Even across this dataset, our model can retrieve images with unseen compositions. As to be expected, there is much more variation. For example, bottle-caps in ImageNet---an object class that is not present in MIT-States---are misconstrued as coins. 

Second, we perform retrieval on the disjoint database \emph{and} issue queries for compositions that are in neither the training nor test set.  For example, the objects \emph{barn} or \emph{cycle} are never seen in MIT-States, under any attribute composition.
We refer to these compositions as \emph{out-of-domain}. 
Our method handles them by applying attribute operators to GloVe object vectors.
\reffig{fig:retrieval-unseen-unseen} (right) shows examples.  
This generalization is straightforward with our method, whereas it is prohibited by the existing methods 
 \SC{RedWine}~\cite{misra2017red} and \SC{AnalogousAttr}~\cite{chen2014inferring}. They rely on having pre-trained SVMs for all constituent concepts. In order to allow an out-of-domain composition with a new object category, those methods would need to gather labeled images for that object, train an SVM, and repeat their full training pipelines.

\vspace*{-0.1in}
\section{Conclusion}

We presented a model of attribute-object composition built around the idea of ``attributes as operators". We modeled this composition as an attribute-conditioned transformation of an object vector, and incorporated it into an embedding learning model to identify unseen compositions. We introduced several linguistically inspired auxiliary loss terms to regularize training, all of which capitalize on the operator model for attributes. Experiments show considerable gains over existing models. 
Our method generalizes well to unseen compositions, in open world, closed world, and even out-of-domain settings. In future work we plan to explore extensions to accommodate relative attribute comparisons and to deal with compositions involving multiple attributes.

\vspace*{0.05in}
{\noindent \textbf{Acknowledgments}: This research is supported in part by ONR PECASE N00014-15-1-2291 and an Amazon AWS Machine Learning Research Award. We gratefully acknowledge Facebook for a GPU donation.}

%
%
%
\bibliographystyle{splncs04}
\bibliography{egbib}

\newpage
\section{Supplementary Details}

This section consists of supplementary material to support the main paper text. The contents include:
\begin{itemize}
	\item Further analysis of the problems with the closed world setting (from Section 4.2 in the main paper) in the context of the MIT-States dataset.
	\item Architecture details of our \SC{LabelEmbed+} baseline model proposed in Section. 4.1 (Baselines) of the main paper.
	\item Variants of baseline models that add our proposed auxiliary regularizer (from Section 3.3).
      \item TSNE visualization of the joint semantic space described in Section 3.1 learned by our method.
	\item Our procedure to obtain the subset of UT-Zappos50K described in Section 4.1 (Datasets) of the main paper.
	\item Additional training details including hyper-parameter selection for all experiments in Section 4.2 of the main paper.

	\item Additional qualitative examples for retrieval on ImageNet from Section 4.3 of the main paper.
	
\end{itemize}

\subsection*{Attribute Affordances: Open vs. Closed world}
As discussed in Section 4.2 of the main paper, recognition in the closed world setting is considerably easier than in the open world setting due to the reduced search space for attribute-object pairs. 

\begin{figure}[t]
\centering
\includegraphics[width=0.5\columnwidth]{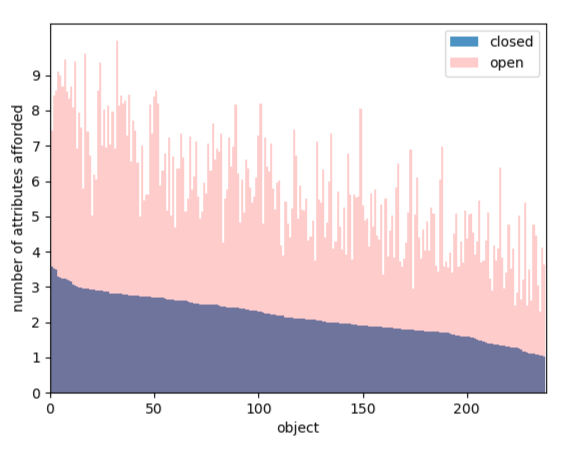}
\caption{\textbf{Attribute affordances for objects}. The closed world setting is easier overall due to the reduced number of attribute choices per object. In addition, about 14\% of the objects are dominated by a single attribute affordance.} 
\label{fig:obj_sparsity}
\end{figure}

\reffig{fig:obj_sparsity} highlights this difference for the MIT-States dataset. In the open world setting, each object would have many potential candidates for compositions (red region), but in the closed world case (blue region), this shrinks to a fraction of compositions. Overall this translates to a 2.8$\times$ higher chance of randomly picking the correct composition in the closed world.  
To make things worse, about 14\% of the objects occur in the test set compositions that are dominated by a single attribute. For example, ``tiger'' affords 2 attributes, but one of those occurs in a single image, leaving \emph{old tiger} as the only relevant composition. A model with poor attribute recognition abilities can still get away with a high accuracy  in terms of the ``closed world" setting as a result, giving a false sense of performance.

Previous work has focused only on the closed world setting, and as a result, compromised on the ability to perform well in the open world (Table 1 in the main text). Our model performs similarly well in both settings, indicating that it does not become dependent on the (artificially) simpler setting of the closed world. 

\subsection*{\SC{LabelEmbed+} Details}

In Section 4.1 (Baselines) of the main paper, we propose the \SC{LabelEmbed+} baseline as an improved baseline model, improving the \SC{LabelEmbed} baseline presented in the \SC{RedWine} paper by Misra et al.  We present the details of the architecture of this baseline here. We use a two layer feed-forward network. Specifically, we concatenate the two primitive input representations of dimension $D$ each, and pass it through a feed-forward network with the configuration (linear-relu-linear) and output dimensions $2D$ and $D$. Unlike \SC{RedWine}, we transform the image representation using a single linear layer, followed by a ReLU non-linearity. We do this to allow some flexibility on the side of the image representation (since we are not finetuning the network responsible for generating the image features themselves).

Here we report additional experiments where we vary the number of layers for this baseline (\reftbl{tab:model_capacity}). We see that our model outperforms \SC{LabelEmbed+} regardless of how many layers are used. This suggests that our improvements are a result of learning a better composition model, and is not related to the network capacity of the baseline model.

\begin{table*}[t]
\centering
\small
\begin{tabular}{l*{3}{S[table-format=3.2]}|*{3}{S[table-format=3.2]}}
                & \multicolumn{3}{c}{MIT-States}  & \multicolumn{3}{c}{UT-Zappos} 	\\ \cline{2-7} 
                & {closed} 	& {open}   & {h-mean} & {closed}  & {open}   & {h-mean} \\ \midrule
LabelEmbed+ (1) & \B{14.9}  & 5.8      & 8.3      & 36.1      & 5.3      & 9.2      \\
LabelEmbed+ (2) & \B{14.9}  & 5.7      & 8.2      & 37.4      & 9.4      & 15.0     \\
LabelEmbed+ (3) & 14.3      & 5.3      & 7.7      & 37.6      & 7.7      & 12.8     \\ \midrule
Ours            & 12.0      & \B{11.4} & \B{11.7} & \B{38.1}  & \B{29.7} & \B{33.4} \\ \bottomrule
\end{tabular}
\caption{\textbf{Model capacity of baseline methods}. The \SC{LabelEmbed+} baseline model with increasing model capacity (number of layers shown in brackets). Our model outperforms this baseline regardless of how many layers are involved, suggesting that model capacity is not the limiting factor.}
\label{tab:model_capacity}
\end{table*}

\subsection*{Baselines Variants}
Next we include modifications to the baselines presented in Section 4.1 of the main paper that are inspired by components of our own model. Specifically, we allow trainable inputs and include the proposed auxiliary regularizer from Section 3.3.

\reftbl{tab:augmented_baselines} shows the results.  
The first column denotes the models as reported in the main paper, while the second column shows the models with extra components from our own model. Note that our model already includes these modifications, and we simply repeat its results on the ``augmented" side for clarity. \SC{RedWine} and \SC{LabelEmbed} are not severely affected because of the way the composition is interpreted---as a set of classifier weights. Extracting the attribute and object identity from \emph{classifier weights} is less meaningful compared to extracting them from a general composition embedding. The auxiliary loss does however improve the embedding learning models. Our model outperforms all augmented variants of the baselines as well.

\begin{table}[t]
\centering
\small
\begin{tabular}{l*{3}{S[table-format=3.2]}|*{3}{S[table-format=3.2]}}
                         & \multicolumn{3}{c}{Original}      & \multicolumn{3}{c}{Augmented Baselines}     \\ \cline{2-7} 
                         & {closed}  & {open}    & {h-mean}  & {closed}  & {open}    & {h-mean} \\ \midrule
\SC{RedWine}             & 12.5      & 3.1       & 5.0       & \B{12.7}  & 3.2       & 5.1      \\
\SC{LabelEmbed}          & 13.4      & 3.3       & 5.3       & 12.2      & 3.1       & 4.9      \\
\SC{LabelEmbed+} 		 & \B{14.9}  & 5.7       & 8.2       & 8.1       & 7.4       & 7.7      \\
\SC{Ours}                & 12.0      & \B{11.4}  & \B{11.7}  & 12.0      & \B{11.4}  & \B{11.7}  \\ \bottomrule
\end{tabular}
\caption{\textbf{Baseline variants on MIT-States}. The proposed 
auxiliary loss term, together with allowing attribute and object representations to be optimized during training, can also help the baselines learn a better composition model.  Our complete model outperforms these baseline variants as well.}
\label{tab:augmented_baselines}
\end{table}

\begin{figure}[t]
\centering
\includegraphics[width=\columnwidth]{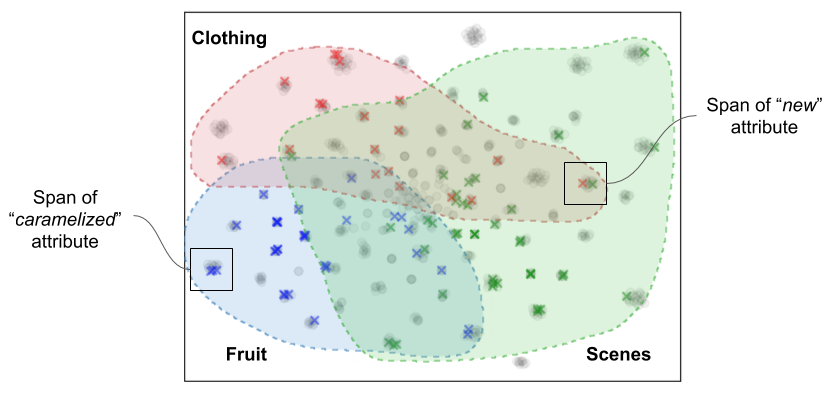}
\caption{\textbf{TSNE visualization of our common embedding space}. 
The span of an attribute operator represents the extent of its application to all objects that afford the attribute.  Here each cluster of black points represents a single attribute's span.
We see a strong separation between compositions in this semantic space, despite these compositions sharing attributes or objects. We highlight composition clusters for three common object superclasses, ``fruit", ``scenes" and ``clothing". Each colored `x' corresponds to a single composition involving an object from the respective superclass. For example, a blue `x' is a composition of the form (attr, fruit) where \emph{attr} is one of (peeled, diced, etc.) and \emph{fruit} is one of (apple, banana, etc.). } 
\label{fig:tsne}
\end{figure}

\subsection*{Visualization of Composition Space}

We visualize the common embedding space described in Section 3.1. \reffig{fig:tsne} contains the 2D TSNE projection of the 300D space generated by our method. The black points represent the embeddings of all unseen compositions in MIT-States. Each cluster (squared) represents the \emph{span} of a single attribute operator---\ie the points in the vector-space that it can reach by transforming object vectors. Our composition model maintains a clear separation between several attribute-object compositions, despite many sharing the same object or attribute. 

We also highlight three object superclasses, ``fruit", ``scenes" and ``clothing", and plot all the compositions they are involved in to show which parts of this subspace are shared among different object classes. We see that common attributes like \emph{old} and \emph{new} are shared by many objects of each superclass, while more specialized attributes like \emph{caramelized} for ``fruit" are separated in this space.

\subsection*{UT-Zappos Subset Selection}

As discussed in Section 4.1 (Datasets) of the main paper, we use a subset of the publicly available UT-Zappos50K dataset in our experiments.  The attributes and annotations typically used in this dataset are \emph{relative attributes}, which are not relevant for our experiments and are not applicable for comparisons to existing work.  However, it also contains labels for \emph{binary material attributes} that are relevant for our experiments.

Here we describe the process for generating the subset of UT-Zappos50K  that we use in our experiments. These images have top level object categories of shoe type (\eg \emph{high heel}, \emph{sandal}, \emph{sneaker}) as well as  finer-grained shoe-type labels (\eg \emph{ankle boots}, \emph{knee-high boots} for the top-level \emph{boots} category). We merge object categories that have fewer than 200 images per category into a single class (\eg all slippers are considered as one class), and discard the sub-classes that do not meet this threshold amount. We then discard all the images that do not have annotations for material attributes of shoes (\eg \emph{leather}, \emph{sheepskin}, \emph{rubber}), which leaves us with $\sim$33K images. We randomly split this set of images into training and testing sets based on their attribute-object compositions. Our subset contains 116 compositions, over 16 attribute classes and 12 object classes.

\subsection*{Additional Training Details}

We provide additional details to accompany Section 4.1 (Implementation Details) in the main paper. 
For our combined loss function, we take a weighted sum of all the losses, and select the weights using a validation set. We create this set for both our datasets by holding out a disjoint subset of 20\% of the training pairs. 

\begin{itemize}
	\item For MIT-States, we train all models for 800 epochs. We set the weight of the auxiliary loss $L_{aux}$ to 1000 for our model.
	\item For UT-Zappos, we train all models for 1000 epochs. We weight all regularizers equally. 
\end{itemize}

The weight for $L_{aux}$ is substantially higher for MIT-States, which may be necessitated by the low volume of training data per composition. On both datasets, we train our models with a learning rate of $1e-4$ and a batch size of 512.

\subsection*{Additional Qualitative Examples}
\begin{figure}[t]
\centering
\includegraphics[width=0.8\columnwidth]{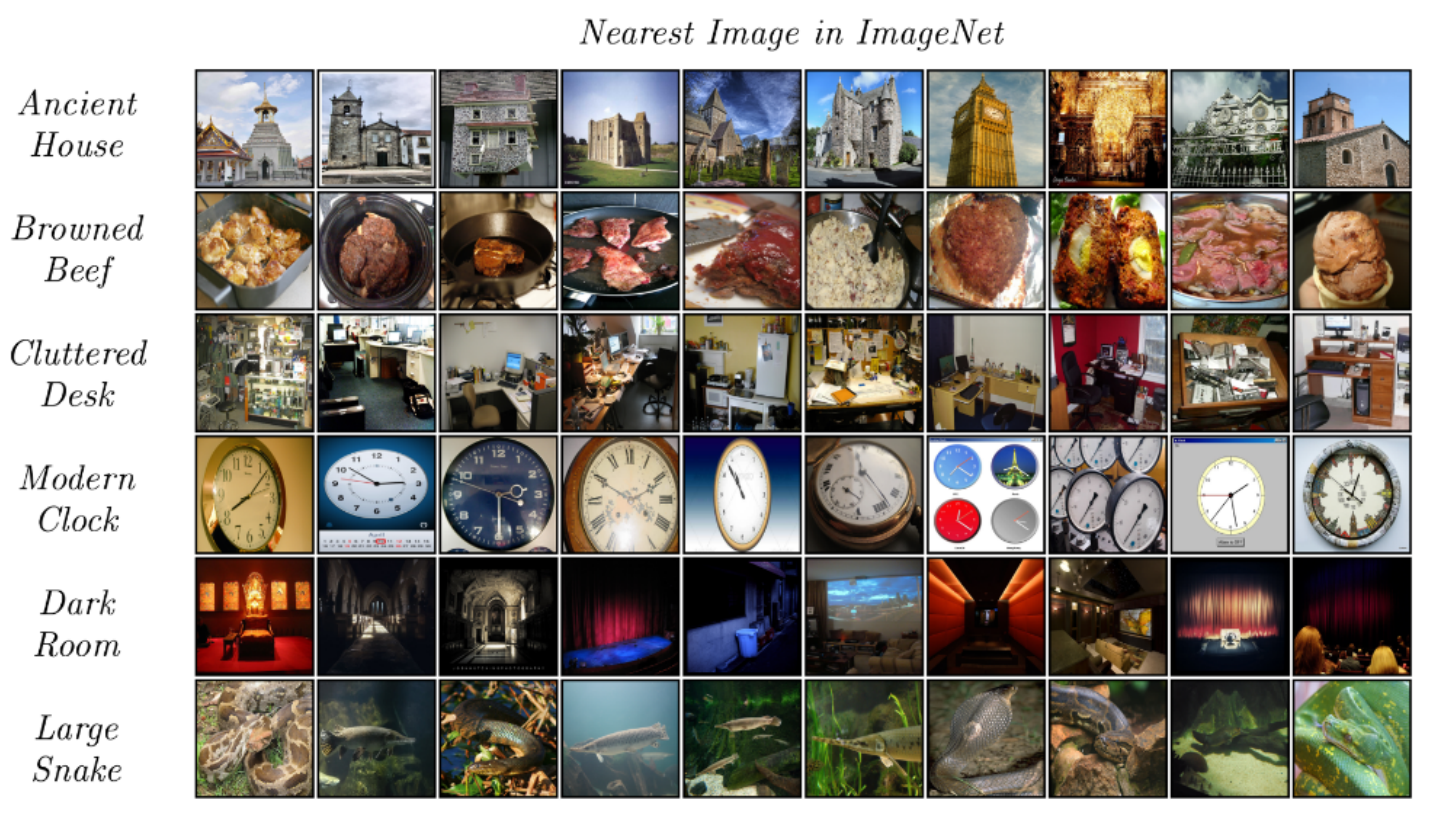}
\caption{\textbf{Retrieval results on ImageNet images}. Text queries of unseen compositions with top-10 image retrievals shown alongside. Note that the compositions are learned from a disjoint set of compositions on a disjoint dataset (MIT-States), then used to issue queries for images in ImageNet.} 
\label{fig:more_imagenet}
\end{figure}

Next we show additional qualitative examples from Section 4.3 for the unseen compositions.  \reffig{fig:more_imagenet} shows retrieval results on a diverse set of images from ImageNet, where object and attribute categories do not directly align with MIT-States.  These examples are computed and displayed in the same manner as Figure 4 in the main paper.

\end{document}